\documentclass[preprint,12pt]{elsarticle}

\usepackage[utf8]{inputenc}
\usepackage[T1]{fontenc}
\usepackage{lmodern}
\usepackage{microtype}

\makeatletter
\def\ps@pprintTitle{%
  \let\@oddhead\@empty
  \let\@evenhead\@empty
  \let\@oddfoot\@empty
  \let\@evenfoot\@empty}
\makeatother

\usepackage{listings}
\usepackage[acronym]{glossaries}
\usepackage{hyperref}
\usepackage{array}
\usepackage{booktabs}
\newacronym{ai}{AI}{Artificial Intelligence}
\newacronym{gi}{GI}{Gastrointestinal}
\newacronym{crc}{CRC}{Colorectal Cancer}
\newacronym{vqa}{VQA}{Visual Question Answering}
\newacronym{qa}{QA}{question--answer}
\newacronym{xai}{XAI}{Explainable Artificial Intelligence}
\newacronym{bleu}{BLEU}{Bilingual Evaluation Understudy}
\newacronym{rouge}{ROUGE}{Recall-Oriented Understudy for Gisting Evaluation}
\newacronym{meteor}{METEOR}{Metric for Evaluation of Translation with Explicit ORdering}

\usepackage[none]{hyphenat}
\begin{document}

\title{Medico 2025: \\ Visual Question Answering for Gastrointestinal Imaging}

\author[1,3]{Sushant Gautam}
\author[1]{Vajira Thambawita}
\author[2]{Michael Riegler}
\author[1,3]{P\aa l Halvorsen}
\author[1]{Steven Hicks}

\address[1]{SimulaMet - Simula Metropolitan Center for Digital Engineering, Oslo, Norway}
\address[2]{Simula Research Laboratory, Oslo, Norway}
\address[3]{OsloMet - Oslo Metropolitan University, Oslo, Norway}

\begin{abstract}
The Medico 2025 challenge addresses \gls{vqa} for \gls{gi} imaging, organized as part of the MediaEval tasks series. The challenge focuses on developing \gls{xai} models that answer clinically relevant questions based on \gls{gi} endoscopy images while providing interpretable justifications aligned with medical reasoning. It introduces two subtasks: (1) answering diverse types of visual questions using the Kvasir-VQA-x1 dataset, and (2) generating multimodal explanations to support clinical decision-making. The Kvasir-VQA-x1 dataset, created from 6,500 images and 159,549 complex \gls{qa} pairs, serves as the benchmark for the challenge. By combining quantitative performance metrics and expert-reviewed explainability assessments, this task aims to advance trustworthy \gls{ai} in medical image analysis. Instructions, data access, and an updated guide for participation are available in the official competition repository: \href{https://github.com/simula/MediaEval-Medico-2025}{github.com/simula/MediaEval-Medico-2025}
\end{abstract}

\maketitle

\section{Introduction}\label{sec:intro}

\acrfull{gi} diseases are among the most common and critical health concerns worldwide, with conditions like \gls{crc} requiring early diagnosis and intervention~\cite{singh2024global, wang2023global}. \gls{ai}-driven decision support systems~\cite{ali2024artificial, berbis2021role} have shown potential in assisting clinicians with diagnosis, but a major challenge remains: explainability. While deep learning models can achieve high diagnostic accuracy, their “black-box” nature limits their adoption in clinical practice, where trust and interpretability are essential~\cite{Borys2023May,Salahuddin2022Jan}.

After successfully organizing multiple Medico challenges at MediaEval in previous years, for the new edition\footnote{\url{https://multimediaeval.github.io/editions/2025/tasks/medico}} we propose the Medico 2025: \textit{\acrlong{vqa} (with multimodal explanations) for \acrlong{gi} Imaging}.

Medical \gls{vqa} is a rapidly growing research area that combines computer vision and natural language processing to answer clinically relevant questions based on medical images~\cite{Borys2023May}. However, existing \gls{vqa} models often lack transparency, making it difficult for healthcare professionals to assess the reliability of \gls{ai}-generated answers~\cite{Borys2023May,Salahuddin2022Jan}. To address this, the Medico 2025 challenge will focus on explainable \gls{vqa} for \gls{gi} imaging, encouraging participants to develop models that provide not only accurate answers but also clear justifications aligned with clinical reasoning.

The challenge provides a benchmark dataset of \gls{gi} images, videos, and associated \gls{vqa} annotations, enabling rigorous evaluation of \gls{ai} models. By integrating multimodal data and explainability metrics, we aim to advance research in interpretable \gls{ai} and increase the potential for clinical adoption.

We define two main subtasks for this year’s challenge. Subtask 2 builds on Subtask 1, meaning Subtask 1 must be completed in order to participate in Subtask 2.

\begin{itemize}
    \item \textbf{Subtask 1:} \gls{ai} Performance on Medical Image Question Answering\\
    This subtask challenges participants to develop \gls{ai} models that accurately interpret and respond to clinical questions based on \gls{gi} images from the Kvasir-VQA-x1 dataset, which retains the original 6,500 images from Kvasir-VQA~\cite{gautam2024kvasir} but expands them to 159,549 \gls{qa} pairs across multiple conditions and instruments. Questions fall into six categories: Yes/No, Single-Choice, Multiple-Choice, Color-Related, Location-Related, and Numerical Count, requiring models to process both visual and textual information. 
    
    Performance will be assessed using \gls{bleu}, \gls{rouge}-1/2/L, and \gls{meteor}.
    \vspace{2cm}
    \item \textbf{Subtask 2:} Clinician-Oriented Multimodal Explanations in \gls{gi}\\ This subtask builds upon Subtask 1, requiring participants to justify their model's predictions using multiple complementary forms of reasoning. The goal is to generate rich, multimodal explanations that are transparent, understandable, and trustworthy to clinicians~\cite{Muhammad2024Dec}. At a minimum, explanations must include a detailed textual narrative in clinical language that directly supports the predicted answer~\cite{Gai2025Apr}. Participants are strongly encouraged to provide an accompanying visual explanation—such as a heatmap, segmentation mask, or bounding box—that clearly links to the textual reasoning and highlights the relevant finding~\cite{Park2018Jun, Storas2025Aug, Dahan2025Jul}. Confidence scores, indicating the model’s certainty, are optional but recommended. 
    
    All outputs will be \emph{human-evaluated} by domain experts and medical professionals, using predefined criteria for clarity, coherence between modalities, and medical relevance, to assess how well the outputs support clinical decision-making.

\end{itemize}

Medical \gls{ai} systems must be both accurate and interpretable to be useful in clinical practice. While deep learning models have shown great potential in diagnosing \gls{gi} conditions from medical images, their adoption remains limited due to a lack of transparency. Clinicians need to understand why an \gls{ai} system makes a specific decision, especially when it comes to critical medical diagnoses. \gls{xai} methods aim to bridge this gap by providing justifications that align with clinical reasoning, improving trust, reliability, and ultimately patient outcomes.

This challenge builds on previous work in medical \gls{vqa}, where \gls{ai} models answer clinically relevant questions based on \gls{gi} images. However, traditional \gls{vqa} models often provide answers without explanations, making it difficult for medical professionals to assess their validity. By incorporating explainability into the task, we encourage the development of models that not only provide accurate responses but also offer meaningful insights into their decision-making process. This will help ensure that \gls{ai} systems can be safely integrated into clinical workflows, assisting rather than replacing human expertise.

\vspace{2cm}
\section{Data}\label{sec:data}

The Medico 2025 challenge builds on the \textbf{Kvasir-VQA-x1} dataset~\cite{Kvasir-VQA-x1}, a substantial extension of the original Kvasir-VQA~\cite{gautam2024kvasir}. It comprises \textbf{6,500} \gls{gi} endoscopic images from HyperKvasir~\cite{borgli2020hyperkvasir} and Kvasir-Instrument~\cite{jha2021kvasir}, paired with \textbf{159,549} \gls{qa} pairs stratified by reasoning complexity.

\begin{table}[!htp]
\centering
\caption{
An example image with one representative \acrlong{qa} pair from each complexity level from the Kvasir-VQA-x1 dataset. Each image in the dataset may have multiple \gls{qa} pairs at every level.
}
\vspace{2mm}
\includegraphics[width=0.5\linewidth]{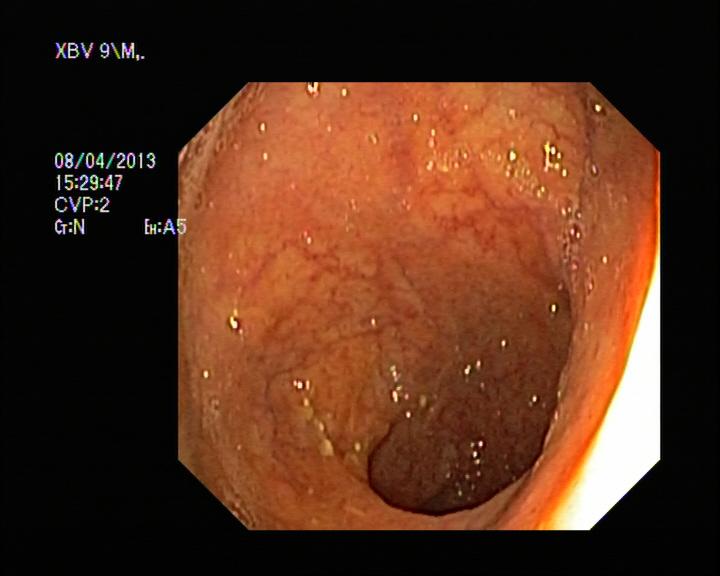}
\vspace{2mm}

\resizebox{\textwidth}{!}{%
\begin{tabular}{@{}c >{\raggedright\arraybackslash}p{6cm} >{\raggedright\arraybackslash}p{6.2cm} >{\raggedright\arraybackslash}p{3.3cm}@{}}
\toprule
\textbf{Complexity} & \textbf{Question} & \textbf{Answer} & \textbf{Question Class} \\
\midrule
1 & Which anatomical landmark is visible in the image? & No identifiable anatomical landmark present & landmark\_location \\
2 & What procedure is depicted in the image and what colors are associated with the abnormality? & Evidence of colonoscopy findings with pink and red mucosal lesions & procedure\_type, abnormality\_color \\
3 & Are there any anatomical landmarks visible, what type of polyps are present, and what colors are the observed abnormalities? & No anatomical landmarks identified, no polyps observed, and multiple abnormalities with pink and red coloration. & landmark\_presence, polyp\_type, abnormality\_color \\
\bottomrule
\end{tabular}
}
\label{tab:example}
\end{table}

\subsection{Dataset Composition}

Each image is paired with multiple \gls{qa} entries generated by merging one to three atomic \gls{qa} pairs using the Qwen3-30B-A3B model~\cite{qwen3technicalreport}. The resulting natural-language questions are fluently phrased, and each entry is annotated with a \texttt{complexity} score (ranging from 1 to 3) and a \texttt{question\_class} label specifying its clinical category. Examples of these classes include \texttt{polyp\_type}, \texttt{instrument\_presence}, and \texttt{finding\_count}.

\subsection{Question Complexity and Clinical Categorization}

The dataset supports stratified evaluation by \gls{qa} complexity:
\begin{itemize}
    \item \textbf{Level 1}: Questions derived from a single atomic \gls{qa} (approximately 34.4\%).
    \item \textbf{Level 2}: Reasoning over two merged atomic \gls{qa} (32.8\%).
    \item \textbf{Level 3}: Synthesis across three atomic \gls{qa} (32.8\%).
\end{itemize}

Each \gls{qa} pair is assigned one or more \texttt{question\_class} labels to support fine-grained analysis across clinical categories such as pathology, anatomical localization, procedural context, and visual findings.

\subsection*{Public Availability and Format}

Kvasir-VQA-x1 is hosted at \url{https://huggingface.co/datasets/SimulaMet/Kvasir-VQA-x1}. The dataset includes \texttt{img\_id} (same as in Kvasir-VQA~\cite{gautam2024kvasir}), \texttt{complexity}, \texttt{question}, \texttt{answer}, \texttt{original} (atomic \gls{qa} components), and \texttt{question\_class}. It is split into training and testing subsets for reproducible experimentation. While only the original images are released, we encourage applying weak augmentations (e.g., rotation, color jitter, crop) when fine-tuning models with the dataset.

\section{Evaluation}

The Medico 2025 challenge evaluates both the \textbf{accuracy} and \textbf{clinical interpretability} of medical \gls{vqa} models, emphasizing not only correct answers but also their \textbf{relevance and explanatory quality} in the context of \gls{gi} diagnostics.

\subsection{Subtask 1: \gls{gi} Question Answering}

This subtask evaluates how effectively models answer clinically relevant \gls{gi} questions from medical images, emphasizing both \textbf{predictive accuracy} and \textbf{reasoning depth}. Performance is measured using language quality metrics—\gls{bleu}, \gls{rouge}-1/2/L, and \gls{meteor}—to assess alignment with reference responses. Evaluation is conducted in two settings: an \emph{original} setting with clean images and a \emph{transformed} setting that applies augmentations for robustness testing. Criteria include accuracy, relevance, and medical correctness.

Evaluation is stratified across three levels: (1) \textit{overall performance}, aggregating scores across all categories and complexities; (2) \textit{category-level analysis} over 18 question types (e.g., polyp type, instrument presence), with visualizations such as radar plots and rank-normalized heatmaps; and (3) \textit{complexity-level evaluation}, distinguishing between factual (Level 1), moderately inferential (Level 2), and higher-order reasoning prompts (Level 3).

This structured, multi-dimensional evaluation framework provides a comprehensive assessment of both correctness and clinical reasoning, which is critical for robust deployment in medical settings.

\subsection{Subtask 2: Clinician-Oriented Explanation Quality Assessment}

This subtask extends Subtask 1 by requiring participants to justify their predictions through detailed, multimodal explanations. The objective is to move beyond providing an answer and produce outputs that are transparent, clinically relevant, and aligned with the model’s own reasoning. 

\textbf{Evaluation in Subtask 2 will be conducted entirely by human experts and medical professionals.} Automated metrics from Subtask 1 (e.g., \gls{bleu}, \gls{rouge}-1/2/L, \gls{meteor}) are used only to assess answer correctness; explanation quality will be judged through expert review according to the criteria below.

Each explanation must combine:
\begin{itemize}
    \item \textbf{Textual Explanation (Mandatory):} A detailed, clinician-oriented narrative that justifies the predicted answer using multiple aspects.
    \item \textbf{Visual Explanation (Optional but Highly Encouraged):} A supporting visual modality—such as a Grad-CAM heatmap, segmentation mask, or bounding box—that highlights the region(s) referenced in the textual explanation. Visuals must clearly link to and reinforce the textual reasoning.
    \item \textbf{Confidence Score (Optional):} A scalar in [0, 1] indicating the model’s certainty in its prediction, derived from softmax probabilities, calibrated uncertainty, or Bayesian methods.
\end{itemize}

Submissions must follow a structured JSON and expert reviewers will rate submissions based on (but not limited to) the following criteria:
\begin{itemize}
    \item \textbf{Clarity:} Ease of understanding for a clinician.
    \item \textbf{Coherence:} Logical consistency between visual and textual components.
    \item \textbf{Medical Relevance:} Consistency with established clinical knowledge.
    \item \textbf{Visual Alignment:} Whether visual elements accurately highlight the relevant findings.
\end{itemize}

By combining accurate predictions with interpretable, clinically grounded justifications, this subtask aims to promote \gls{ai} systems that can be meaningfully integrated into real-world diagnostic workflows.

\section{Discussion and Outlook}

The Medico 2025 challenge marks an important step toward bridging the gap between powerful deep learning models and their practical adoption in clinical settings. By focusing on explainable \gls{vqa} for \gls{gi} imaging, this task promotes the development of interpretable \gls{ai} models that not only generate accurate responses but also provide transparent justifications aligned with medical reasoning.

Participants are encouraged to innovate beyond traditional accuracy metrics and embrace multimodal explainability as a core component of their solutions. The availability of the Kvasir-VQA-x1 dataset, tailored for this task, will support reproducible research and enable robust benchmarking.

Looking ahead, we anticipate that methods developed for Medico 2025 will inspire broader applications of explainable \gls{ai} in other medical domains. By fostering interdisciplinary collaboration between the \gls{ai} and medical communities, this challenge aims to pave the way for clinically viable \gls{ai} tools that are both trusted and actionable in real-world healthcare scenarios.

\def\bibfont{\small} %
\bibliographystyle{plainnat}
\bibliography{references} 

\end{document}